\documentclass[conference,a4paper,english]{IEEEtran}[2015/08/26]

\usepackage{balance}

\usepackage{upquote}

\usepackage[ngerman,main=english]{babel}
\addto\extrasenglish{\languageshorthands{ngerman}\useshorthands{"}}

\usepackage[hyphens]{url}

\makeatletter
\g@addto@macro{\UrlBreaks}{\UrlOrds}
\makeatother

\usepackage[zerostyle=b,scaled=.75]{newtxtt}

\usepackage[T1]{fontenc}

\usepackage[
  babel=true, 
  expansion=alltext,
  protrusion=alltext-nott, 
  final 
]{microtype}

\DisableLigatures{encoding = T1, family = tt* }

\usepackage{graphicx}

\usepackage{diagbox}

\usepackage{xcolor}

\usepackage{listings}

\usepackage{graphicx}
\usepackage{amsmath}
\usepackage{amssymb}
\usepackage{booktabs}
\usepackage{makecell}

\usepackage{pifont}
\usepackage{lipsum}
\usepackage{hhline}
\usepackage{multirow}
\usepackage{tabularx}
\definecolor{eclipseStrings}{RGB}{42,0.0,255}
\definecolor{eclipseKeywords}{RGB}{127,0,85}
\colorlet{numb}{magenta!60!black}

\lstdefinelanguage{json}{
    basicstyle=\normalfont\ttfamily,
    commentstyle=\color{eclipseStrings}, 
    stringstyle=\color{eclipseKeywords}, 
    numbers=left,
    numberstyle=\scriptsize,
    stepnumber=1,
    numbersep=8pt,
    showstringspaces=false,
    breaklines=true,
    frame=lines,
    string=[s]{"}{"},
    comment=[l]{:\ "},
    morecomment=[l]{:"},
    literate=
        *{0}{{{\color{numb}0}}}{1}
         {1}{{{\color{numb}1}}}{1}
         {2}{{{\color{numb}2}}}{1}
         {3}{{{\color{numb}3}}}{1}
         {4}{{{\color{numb}4}}}{1}
         {5}{{{\color{numb}5}}}{1}
         {6}{{{\color{numb}6}}}{1}
         {7}{{{\color{numb}7}}}{1}
         {8}{{{\color{numb}8}}}{1}
         {9}{{{\color{numb}9}}}{1}
}

\usepackage[autostyle=true]{csquotes}

\defineshorthand{"`}{\openautoquote}
\defineshorthand{"'}{\closeautoquote}

\usepackage{booktabs}

\usepackage{paralist}

\usepackage[%
  square,        
  comma,         
  numbers,       
  sort&compress  
]{natbib}

\usepackage{etoolbox}
\makeatletter
\patchcmd{\NAT@test}{\else \NAT@nm}{\else \NAT@hyper@{\NAT@nm}}{}{}
\makeatother

\usepackage{pdfcomment}

\newcommand\blfootnote[1]{%
  \begingroup
  \renewcommand\thefootnote{}\footnote{#1}%
  \addtocounter{footnote}{-1}%
  \endgroup
}
\usepackage{stfloats}
\fnbelowfloat

\usepackage[group-minimum-digits=4,per-mode=fraction]{siunitx}

\usepackage{hyperref}
\hypersetup{
  hidelinks,
  colorlinks=true,
  allcolors=black,
  pdfstartview=Fit,
  breaklinks=true
}

\usepackage[all]{hypcap}

\usepackage[caption=false,font=footnotesize]{subfig}

\usepackage[incolumn]{mindflow}
\usepackage[capitalize]{cleveref}

\crefname{listing}{Listing}{Listings}
\Crefname{listing}{Listing}{Listings}
\crefname{lstlisting}{Listing}{Listings}
\Crefname{lstlisting}{Listing}{Listings}

\usepackage{lipsum}

\usepackage[math]{blindtext}
\usepackage{mwe}
\usepackage[realmainfile]{currfile}
\usepackage{tcolorbox}
\tcbuselibrary{listings}

\DeclareFontFamily{U}{MnSymbolC}{}
\DeclareSymbolFont{MnSyC}{U}{MnSymbolC}{m}{n}
\DeclareFontShape{U}{MnSymbolC}{m}{n}{
  <-6>    MnSymbolC5
  <6-7>   MnSymbolC6
  <7-8>   MnSymbolC7
  <8-9>   MnSymbolC8
  <9-10>  MnSymbolC9
  <10-12> MnSymbolC10
  <12->   MnSymbolC12%
}{}
\DeclareMathSymbol{\powerset}{\mathord}{MnSyC}{180}

\usepackage{xspace}

\newcommand{\ie}{i.e.,\ }

\makeatletter
\newcommand{\hydash}{\penalty\@M-\hskip\z@skip}
\makeatother

\input glyphtounicode
\begin{document}

\title{Efficient Coarse-to-Fine Diffusion Models \\
with Time Step Sequence Redistribution\\\vspace*{-0.5cm}}

\author{%
  \IEEEauthorblockN{Yu-Shan Tai}
  \IEEEauthorblockA{\textit{Graduate Institute of Electrical Engineering}\\\textit{National Taiwan University}\\Taipei, Taiwan\\
    clover@access.ee.ntu.edu.tw\\\vspace*{-1.3cm}}
  \and
  \IEEEauthorblockN{An-Yeu (Andy) Wu}
  \IEEEauthorblockA{\textit{Graduate Institute of Electrical Engineering}\\\textit{National Taiwan University}\\Taipei, Taiwan\\
    andywu@ntu.edu.tw\\\vspace*{-1.3cm}}
}

\maketitle
%
%
\begin{abstract}
Recently, diffusion models (DMs) have made significant strides in high-quality image generation. 
However, the multi-step denoising process often results in considerable computational overhead, impeding deployment on resource-constrained edge devices. 
Existing methods mitigate this issue by compressing models and adjusting the time step sequence.
However, they overlook input redundancy and require lengthy search times.
In this paper, we propose Coarse-to-Fine Diffusion Models with Time Step Sequence Redistribution. 
Recognizing indistinguishable early-stage generated images, we introduce Coarse-to-Fine Denoising (C2F) to reduce computation during coarse feature generation. 
Furthermore, we design Time Step Sequence Redistribution (TRD) for efficient sampling trajectory adjustment, requiring less than 10 minutes for search. 
Experimental results demonstrate that the proposed methods achieve near-lossless performance with an 80\% to 90\% reduction in computation on CIFAR10 and LSUN-Church.
\blfootnote{This work is financially supported by National Science and Technology Council, Taiwan, under grants NSTC 113-2218-E-002-036-MBK and NSTC 113-2221-E-002-120-MY3. }
\end{abstract}

\vspace{-0.7em}
%
\IEEEpeerreviewmaketitle

\section{Introduction}
\label{sec:intro}

Diffusion models (DMs) have showcased remarkable progress in image generation \cite{ddpm, denoising, improved, dm_high, dm_lugmayr2022repaint, dm_srdiff, dm_whang2022deblurring}, video generation \cite{video, video_imagen, videoDM}, and 3D point cloud generation \cite{point_cloud, 3d_conditional, 3d_difffacto, 3d_holodiffusion}. 
These models gradually introduce noise into input data until it conforms to a Gaussian distribution, then learn to denoise and restore real data.
Despite their impressive achievements, DMs suffer from high computational demand.
Unlike GANs \cite{gan}, which require a single forward pass to generate a sample, DMs involve hundreds to thousands of iterative steps. 
This high demand for computation hinders their deployment on resource-constrained edge devices.
Therefore, minimizing the computational burden of the denoising process is essential, requiring immediate strategies to balance image quality with computational efficiency.

To address this challenge, researchers have proposed various approaches to improve the denoising process.
Some aim to reduce the model overhead per step.
For instance, Diff-Pruning \cite{diff-pruning} leverages Taylor expansion over pruned time steps to identify and eliminate non-critical parameters. 
Some others propose advanced schedulers that adjust sampling trajectories from a numerical perspective \cite{ddim, dpm_solver++}. 
However, these approaches degrade image quality as step numbers decrease. 
To address this, AutoDiffusion \cite{autodiffusion} further applies evolutionary search to find the optimal time step sequence.

\begin{figure}[t]
    \centering
    \includegraphics[width=0.9\columnwidth]{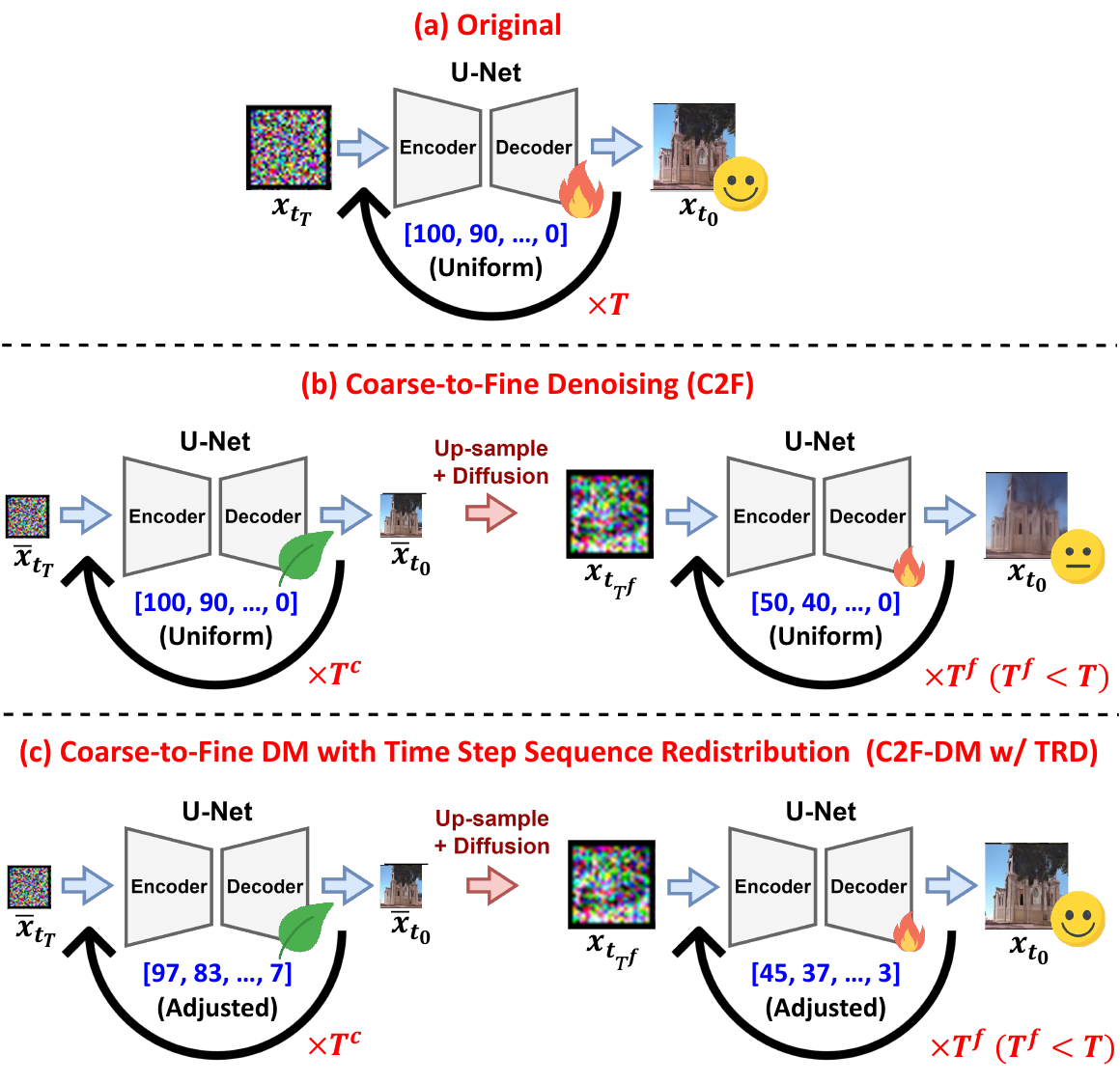}
    \vspace{-1em}
    \caption{(a) Original model, (b) C2F, and (c) C2F-DM w/ TRD.}
   \label{fig:overview}
    \vspace{-2em}
\end{figure}
        
Though the aforementioned methods offer solutions to certain challenges, there are still issues to be addressed:
\begin{enumerate}
    \item \textit{Unnecessary computation during denoising}: 
        The computation of DMs scales quadratically with image resolution, incurring huge costs for large images.
        However, some steps do not require such intensive computation.
    \item \textit{Time-consuming search for time step sequence}: 
        While effective in enhancing image quality with limited time steps, evolutionary search \cite{autodiffusion} takes more than one day due to the time-consuming evaluation process.
\end{enumerate}
To address these issues, we propose Coarse-to-Fine Diffusion Models with Time Step Sequence Redistribution, as shown in \cref{fig:overview}. Our main contributions are as follows: 
\begin{enumerate}
    \item \textit{Coarse-to-Fine Denoising (C2F)}: 
        We propose starting with low-resolution denoising and transitioning to high-resolution denoising later, reducing computation during coarse feature generation.

    \item \textit{Time Step Sequence Redistribution (TRD)}: 
        We use iterative substitution to efficiently adjust time step sequences, preserving image quality in fewer steps with an offline search time under 10 minutes.
    \item
        We validated on CIFAR10 \cite{cifar} and LSUN-Church \cite{lsun} with different schedulers, DDIM \cite{ddim} and DPM-Solver++ \cite{dpm_solver++}. Our results exhibit near-lossless performance with an 80\% to 90\% reduction in computation.
\end{enumerate}  


\section{Related works}\label{sec:related_work}
    \subsection{Diffusion Models (DMs)}\label{ssec:DM}
    Diffusion models generate images using a Markov chain. 
    Given a real data distribution $x_{t_0} \sim q(x_{t_0})$, the diffusion process gradually adds Gaussian noise with a predefined variance schedule $\beta_{t_i} \in (0,1)$ to produce an intermediate input sequence $x_{t_1}, ..., x_{t_T}$ for $T$ steps.
    When $t_T\rightarrow\infty$, $x_{t_T}$ approaches a Gaussian distribution.
    \begin{equation}
        q(x_{t_i}|x_{t_{i-1}})=\mathcal{N}(x_{t_i};\sqrt{1-\beta_{t_i}}x_{t_{i-1}},\beta_{t_i}\textbf{I}).
        \label{eq:diffusion}
    \end{equation}
    Conversely, the denoising process aims to remove noise from a noise input $x_{t_T}\in \mathcal{N}(\textbf{0},\textbf{I})$ to generate the final output $x_{t_0}$.
    \begin{equation}
        p_\theta(x_{t_{i-1}}|x_{t_{i}})=\mathcal{N}(x_{t_{i-1}};\tilde{\mu}_{\theta,t_i}(x_{t_i}),\tilde{\beta}_{t_i}\textbf{I}).
        \label{eq:diffusion}
    \end{equation}
    A noise estimation model parameterized by $\theta$ is applied to deduce the inputs for the next step, $x_{t_{i-1}}$, and the current prediction of the final images, $\hat{x}_{t_0}^i$.
    The noise estimation model is typically a U-Net \cite{unet}, a convolutional neural network whose computation grows quadratically with image resolution.

    \begin{figure}[t]
        \centering
        \includegraphics[width=0.9\columnwidth]{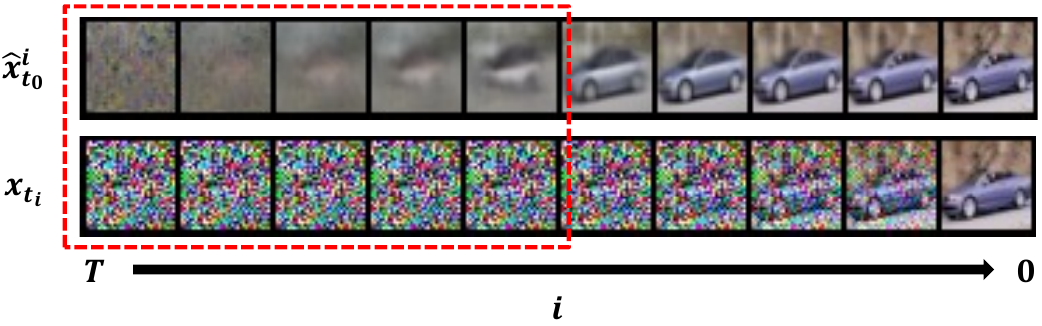}
        \vspace{-1em}
        \caption{Prediction of the final images $\hat{x}_{t_0}^i$ and intermediate inputs $x_{t_i}$ during denoising, which are noisy in the early stage.}
       \label{fig:denoise}
    \vspace{-1em}
    \end{figure}
    
    \subsection{Computation Reduction for Diffusion Models}\label{ssec:Acceleration}
    \subsubsection{Optimization for Per Step Overhead}\label{sssec:per step}
    To alleviate the computational burdens, various strategies have been proposed to optimize the denoising process in DMs.
    Some researchers aim to reduce the model overhead per step by deploying a light U-Net.
    For instance, Diff-pruning \cite{diff-pruning} employs Taylor expansion over pruned time steps to assess the importance of parameters and eliminate non-critical ones. 
    Additionally, quantization techniques \cite{ptq4dm,q_diffusion} select appropriate calibration data to obtain low-precision models.
    However, unlike these approaches \cite{diff-pruning, ptq4dm, q_diffusion} that focus on model compression, we observed that simplifying inputs can also alleviate computational burdens. 
    Consequently, we propose Coarse-to-Fine Denoising (C2F) to reduce computation overhead for coarse feature generation.
    
    \subsubsection{Reduction of Denoising Steps}\label{sssec:per step}
    Some other works pursue simulating the denoising process in fewer steps while maintaining performance.
    DDIM \cite{ddim} employs a non-Markovian process, using a uniform sub-sequence of DDPM \cite{ddpm} to reduce steps. 
    Furthermore, DPM-Solver++ \cite{dpm_solver++} explores the fast solver of ordinary differential equations (ODEs) to create efficient sampling.
    Unfortunately, these training-free schedulers \cite{ddim, dpm_solver++} suffer serious performance degradation when the number of steps is below 10.
    To address this issue and enhance the image quality within limited steps, AutoDiffusion \cite{autodiffusion} proposes a neural architecture search (NAS) system for DMs, utilizing evolutionary search to find the optimal time step sequence. 
    However, the search process requires more than one day due to the iterative and time-consuming evaluation of each candidate.
    To enhance the efficiency, we introduce Time Step Sequence Redistribution (TRD), which takes less than 10 minutes.  
    
\section{Proposed Methods}\label{sec:method}
    \subsection{Coarse-to-Fine Denoising (C2F)}\label{ssec:C2F}
    \subsubsection{Overall Flow}\label{sssec:C2F_overall}
    Given that the computation of DMs increases quadratically with image resolution, generating large images demands substantial computation resources.
    Hence, it would be advantageous to utilize lower-resolution (\ie\space smaller-sized) inputs to replace some intensive computations.
    Inspired by this insight, we examine the prediction of the final images $\hat{x}_{t_0}^i$ and intermediate inputs $x_{t_i}$ during denoising, as shown in \cref{fig:denoise}. 
    In the early stages (enclosed by the red dashed line), $\hat{x}_{t_0}^i$ appear notably noisy and indistinguishable, with only coarse features like structure and shape being slightly recognizable.
    It is only towards the later stages of denoising that the finer details begin to emerge.

    Based on our observation, we introduce the Coarse-to-Fine Denoising (C2F), as depicted in \cref{fig:overview}(b). 
    Firstly, we perform low-resolution denoising with $T^c$ steps to obtain the noiseless images, $\bar{x}_{t_{0}}$.
    Then, we up-sample $\bar{x}_{t_{0}}$ to the original size and deduce the high-resolution intermediate inputs $x_{t_{T^f}}$.
    Finally, the denoising process operates under high resolution for $T^f$ steps to produce the final images, $x_{t_{0}}$. 
    
    Since the computation of DMs scales quadratically with resolution, processing low-resolution inputs is significantly faster than running high-resolution ones.
    Thus, the proposed C2F efficiently reduces computation for generating coarse features, easing the computational load during denoising.
    
    \subsubsection{Multi-resolution Fine-tuning}\label{sssec:C2F_multi}
\begin{figure}[t]
        \centering
        \includegraphics[width=0.9\columnwidth]{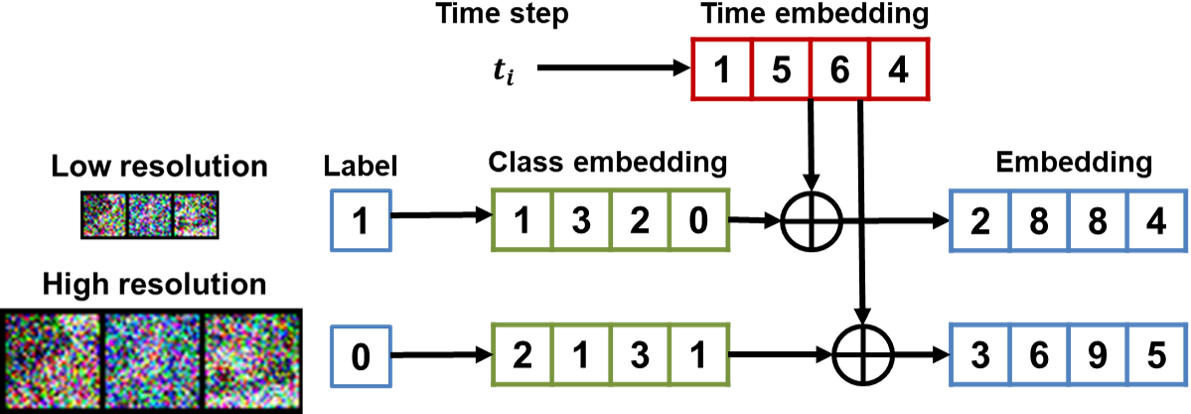}
        \vspace{-1em}
        \caption{Overview of multi-resolution fine-tuning .}
       \label{fig:multi}
    \vspace{-1.5em}
    \end{figure}

    While the inputs of DMs can vary in size due to the convolution property, generated images may become corrupted if the input size differs from the original training size.
    Instead of employing multiple DMs to handle each resolution independently, we propose multi-resolution fine-tuning to process both low and high resolutions through a single pre-trained DM.
    To enable DM to recognize inputs at varying resolutions, we assign different labels to the two resolutions, transform them into class embeddings, and then add them with time embeddings, as illustrated in \cref{fig:multi}. 
    Then, we alternatively apply low-resolution and high-resolution inputs to fine-tune the embeddings and weights of a pre-trained DM.
    This strategy successfully facilitates DMs in generalizing to different resolutions.
    Additionally, we explore other multi-resolution fine-tuning strategies, as detailed in \cref{ssec:multi-resolution_compare}.

    \subsubsection{Initial Step for High-resolution Denoising}\label{sssec:C2F_PCA}
\begin{figure}[t]
        \centering
        \vspace{-0.5em}
        \includegraphics[width=0.8\columnwidth]{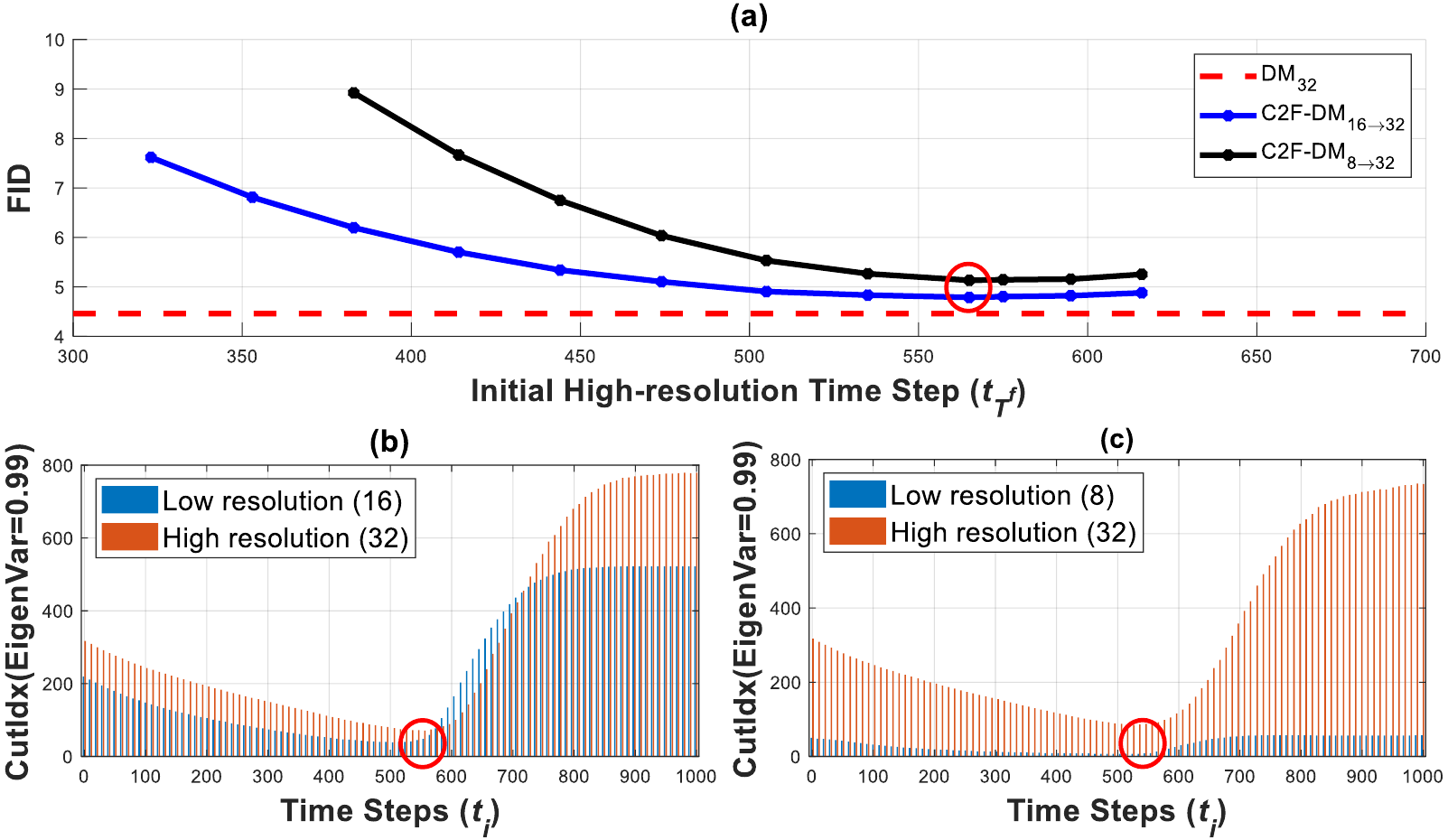}
        \vspace{-1em}
        \caption{(a) FID ($\downarrow$) under different $t_{T^f}$ and the cut index of Eigen variance 0.99 on $\hat{x}^i_{t_0}$ of (b) C2F-DM$_{16\rightarrow32}$ and (c) C2F-DM$_{8\rightarrow32}$.
        DM$_{32}$ denotes the original DM.
        For DMs employing C2F, with the low resolution set at 16 and 8, we denote them as C2F-DM$_{16\rightarrow32}$ and C2F-DM$_{8\rightarrow32}$.}
       \label{fig:PCA}
        \vspace{-1.8em}
    \end{figure}

    After acquiring the DM applicable to both low and high resolutions, our next objective is to determine the optimal initial step $t_{T^f}$ for high-resolution denoising. 
    The Frechet Inception Distance (FID) \cite{fid} on CIFAR10 (32$\times$32) under varying $t_{T^f}$ is shown in \cref{fig:PCA}(a).
    Remarkably, FID tends to increase as $t_{T^f}$ decreases.
    This is reasonable as a smaller $t_{T^f}$ implies fewer steps to operate under high resolution, thereby compromising image quality.
    Thus, the optimal choice would be the smallest $t_{T^f}$ yielding the lowest FID, approximately at 565 in the case. 
    To further analyze, we conduct Principal Component Analysis (PCA) on the prediction of final images $\hat{x}_{t_0}^i$ during denoising, as shown in \cref{fig:PCA}(b) and (c).
    The cut index serves as an indicator of ranks.
    Surprisingly, we observe three interesting properties:
    (1) Generally, higher resolutions correspond to higher ranks. (2) Ranks initially decrease and then enhance during the denoising process. (3) The optimal $t_{T^f}$ identified in \cref{fig:PCA}(a) coincides with $t_{i}$  associated with the minimal ranks for low-resolution images in \cref{fig:PCA}(b) and (c), as marked by the red circles.
    These observations align with different stages of the denoising process, as shown in \cref{fig:denoise}.
    Initially, DMs focus on removing noise, leading to a decrease in rank. Then, it reaches a minimum rank point and starts generating details,  leading to an increase in rank. 
    Assigning $t_{T^f}$ to the transition point of removing noise and generating details is justifiable, as finer features are better preserved under high resolution.
    This analysis aids us in efficiently determining the optimal $t_{T^f}$ without time-consuming evaluations at each time step.
    \begin{figure*}[t]
        \centering
        \includegraphics[width=1.8\columnwidth]{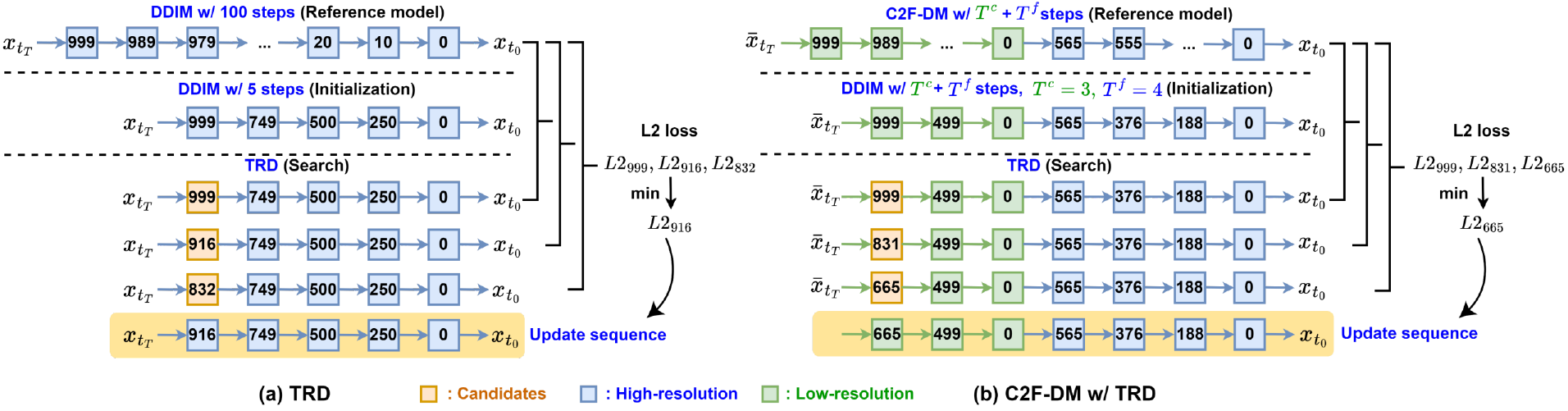}
        \vspace{-1em}
        \caption{(a) Time Step Sequence Redistribution (TRD). (b) Combined flow (C2F-DM w/ TRD).}
       \label{fig:TRD}
        \vspace{-1.5em}
    \end{figure*}
    
        \vspace{-1.5em}

    \vspace{1em}
    
    \subsection{Time Step Sequence Redistribution (TRD)}\label{ssec:TRD}
    After reducing the computation for coarse features by C2F, we further introduce the Time Step Sequence Redistribution (TRD) method to refine the sampling trajectories.
    The workflow of TRD is depicted in \cref{fig:TRD}(a), where the objective is to condense the number of steps from 100 to 5, using DDIM \cite{ddim} with 100 steps as the reference model.
    We begin with the standard uniform time step sequences and progressively substitute each step if its nearby candidates are considered more suitable.
    The candidate sets $C_{i}$ comprise $n$ linearly divided steps between the current step $t_i$ and its neighbors:
    \begin{equation}
        C_{i} = \{t_{i-1}+j\cdot \frac{t_{i+1}-t_{i-1}}{n+1}|j=[1,n],0<i<T\}.
        \label{eq:candidates}
    \end{equation}
    For $t_T$ and $t_0$, we utilize linearly divided steps with their sole neighbor as candidates.
    Then, we compute the L2 loss between $x_{t_0}$ generated by each candidate and the reference model with the calibration set.
    The candidate with the minimum L2 loss is deemed the most appropriate and used to replace the original time step. 
    This substitution process iterates for each step until no further updates occur or the maximum iteration is reached.

    Instead of generating numerous samples for FID evaluation \cite{autodiffusion}, we only generate 16 images to calculate L2 loss and focus solely on nearby candidates. This approach significantly speeds up the search process, allowing us to derive an improved time step sequence in less than 10 minutes. In \cref{ssec:TRD_calib}, we will provide a detailed analysis to showcase its efficiency.

    \subsection{Combined Flow (C2F-DM w/ TRD)}\label{ssec:combine}
    To pursue the combined benefits of C2F and TRD, we further integrate them, as shown in \cref{fig:overview}(c). 
    However, this fusion required some modifications to ensure feasibility, as shown in \cref{fig:TRD}(b).
    Firstly, the original DM cannot be the reference model because the generated images depend on the inputs. 
    For accurate comparison of the final images $x_{t_0}$, the reference model and candidates must have the same inputs. Unfortunately, since the input size of C2F-DM is smaller than that of the original DM, they generate different images. Thus, we utilize the optimal C2F-DM identified through PCA, as discussed in \cref{sssec:C2F_PCA}, to serve as the reference model.
    Secondly, we should determine the initialization of the time step sequence. 
    For low-resolution denoising, we apply the uniformly divided $T^c$ steps among $[t_T,0]$.
    However, this rule cannot apply to high-resolution denoising, as starting from $t_T$ means re-sampling a noise input, discarding all features generated under low resolution.
    Hence, we employ uniformly divided $T^f$ steps among $[t^*,0]$ for the initialization of high-resolution steps, where $t^*$ is the optimal initial high-resolution time steps found in \cref{sssec:C2F_PCA}.
    After that, we follow the same procedure mentioned in \cref{ssec:TRD} to traverse each step to optimize the time step sequence.

    The combined flow incorporates both the advantages of C2F and TRD, significantly improving the efficiency and quality of the denoising process.
    In \cref{ssec:results}, we would delve into the qualitative results to validate the effectiveness.
    
\section{Simulation Results}
    We experiment on unconditional generation using CIFAR10 (32$\times$32) \cite{cifar} and LSUN-Church (256$\times$256) \cite{lsun} 
    , focusing on DDPMs \cite{ddpm}.
    For C2F, we fine-tune 100k and 375k iterations on CIFAR10 and LSUN-Church, less than 10\% of the standard training.
    If unspecified, we set the low resolution to the smallest input size compatible with DMs, 
    \ie 8 for CIFAR10 and 32 for LSUN-Church.
    We set the calibration size to 16, candidate number $n$ to 5, and maximum iteration to 10.

    \begin{figure}[t]
        \centering
        \includegraphics[width=0.9\columnwidth]{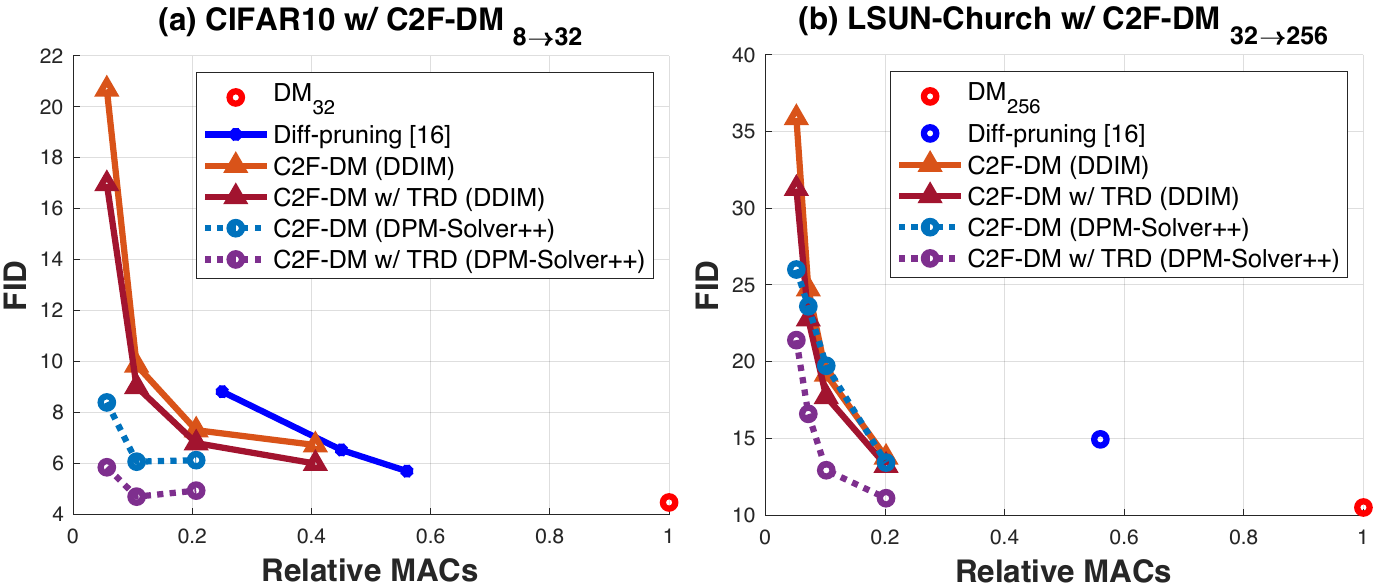}
        \vspace{-1em}
        \caption{MACs ($\downarrow$) and FID ($\downarrow$) on (a) CIFAR10 and (b) LSUN-Church.}
       \label{fig:fid}
        \vspace{-1em}
    \end{figure}
    
    \begin{figure}[t]
        \centering
        \includegraphics[width=0.9\columnwidth]{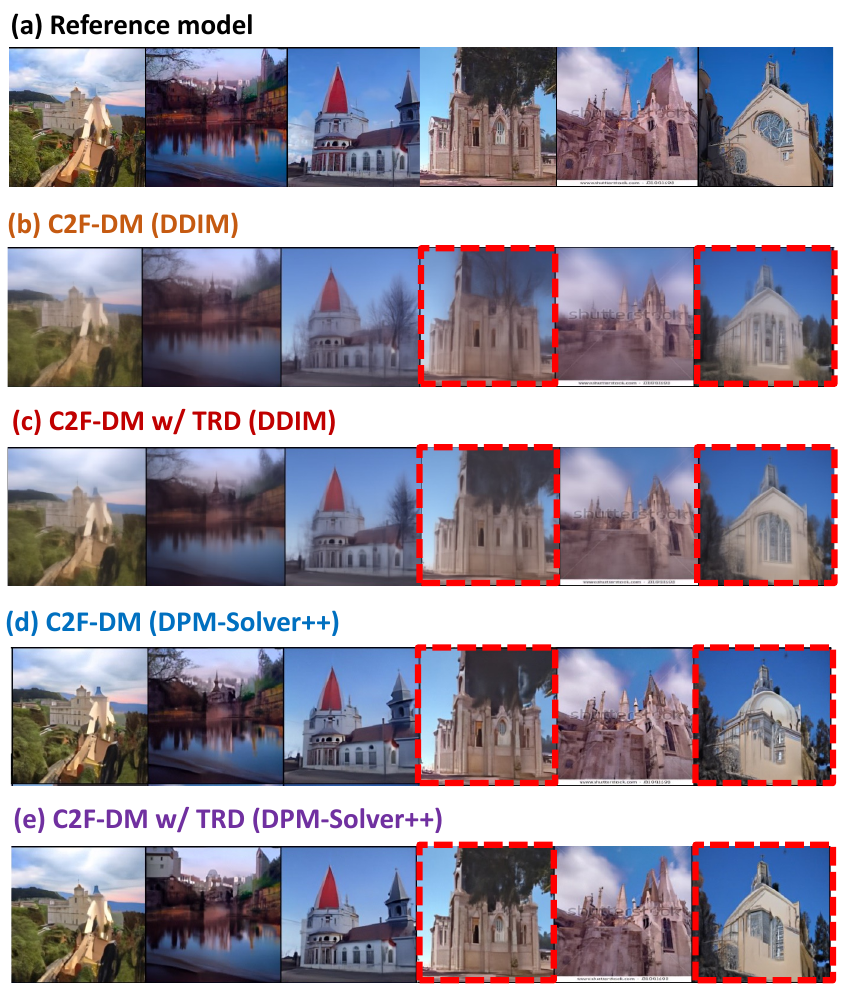}
        \vspace{-1em}
        \caption{Generated images $x_{t_0}$ under $T^f=5$ for LSUN-Church. 
        }
       \label{fig:visual}
        \vspace{-1em}
    \end{figure}
    
    \subsection{Qualitative Results} \label{ssec:results}
    In this section, we compare our method to Diff-pruning \cite{diff-pruning} by the relative MACs to 100-step DDIM and FID, as depicted in \cref{fig:fid}.
    For CIFAR10 and LSUN-Church, the numbers of high-resolution steps $T^f$ are [5,10,20,40] and [5,7,10,20], respectively, while the number of low-resolution steps $T^c$ is set to 10 for both datasets.
    Compared with \cite{diff-pruning}, \textit{C2F-DM (DDIM)} achieves lower MACs with similar FID, indicating notable efficiency gains by using light computations during coarse feature generation.
    Moreover, with the inclusion of TRD, \textit{C2F-DM w/ TRD (DDIM)} achieves improved FID under the same number of steps.
    Additionally, we evaluate the generalizability of our method by applying it to another scheduler, DPM-Solver++ \cite{dpm_solver++}. Similar to the DDIM case, the results show that \textit{C2F-DM w/ TRD (DPM-Solver++)} outperforms the counterpart without TRD, highlighting its effectiveness in selecting essential steps. 
    Notably, we achieve near-lossless performance with a 90\% reduction in MACs on CIFAR10 and an 80\% reduction on LSUN-Church.
    
    In \cref{fig:visual}, we display the generated images under $T^f=5$, corresponding to the leftmost point of each line in \cref{fig:fid}(b). The colors of the titles match those of the lines.
    Remarkably, images generated with TRD exhibit clearer details and better preserve intricate features, such as tree leaves and windows.

    
    \subsection{Different Multi-resolution Fine-tuning Strategies} \label{ssec:multi-resolution_compare}
    In \cref{sssec:C2F_multi}, we introduce a multi-resolution fine-tuning strategy. Here, we further compare with others strategies: (1) \textit{1DM-xy}: We concatenate normalized (\textit{x,y}) coordinates with inputs. (2) \textit{1DM-2T}: We modify the time step sequence to contain $2T$ steps, where $2 \cdot t_T \sim t_0+t_T$ for low resolution and $t_T \sim t_0$ for high resolution. (3) \textit{1DM-label}: The strategy used in \cref{sssec:C2F_multi}.
    Additionally, we also compare with the inefficient counterpart, \textit{2DMs}, which uses two individual models for high and low resolutions.
    The FID of C2F-DM$_{16\rightarrow32}$ under different strategies on CIFAR10 are presented in \cref{tab:multi}. 
    Among the three strategies, \textit{1DM-label} achieves the lowest FID, close to the performance of \textit{2DMs}.
    Consequently, we select \textit{1DM-label} as our final multi-resolution fine-tuning strategy.

    \begin{table}[t]
        \centering
        \caption{FID of different multi-resolution fine-tuning strategies}
        \vspace{-1em}
        \resizebox{0.47\textwidth}{!}{
            \begin{tabular}{c|cc|ccc}
            \Xhline{1pt}
            Method          & Original   & \textit{2DMs}      & \textit{1DM-xy}      & \textit{1DM-2T}   & \textit{1DM-label}       \\ \hhline{======}
            FID $\downarrow$   & 4.46 & 4.70 & 5.48       & 6.03      & \textbf{4.91}         \\ \hline
            \end{tabular}
            }
            \label{tab:multi}
        \vspace{-1em}
        \end{table}
    
    \subsection{Search Time Analysis} \label{ssec:TRD_calib} 
    \begin{table}[t]
        \centering
        \caption{Search time under different datasets ($T^f=5$)}
        \vspace{-1em}
        \resizebox{0.4\textwidth}{!}{
            \begin{tabular}{c|cc}
            \Xhline{1pt}
            Dataset          & CIFAR10   & LSUN-Church    \\ \hhline{===}
            Search time (minutes)   & 1.4 & 8.8         \\ \hline
            \end{tabular}
            }
            \label{tab:time}
        \vspace{-2em}
        \end{table}
    

    To ensure the practicality of our proposed approach, we delve into the time analysis of calibration.
    In \cref{tab:time}, we demonstrate the search time required under $T^f=5$ on NVIDIA GeForce RTX 4090.
    Using a small calibration set and evaluating by L2 loss instead of FID, the heuristic process can be efficiently completed in less than 10 minutes.
    This rapid calibration highlights the practicality and efficiency of our method.


\section{Conclusion}
    In this paper, we introduce Coarse-to-Fine Diffusion Models with Time Step Sequence Redistribution. Specifically, C2F saves computation for coarse features while TRD efficiently identifies an enhanced sampling trajectory. 
    Experimental results show our method reaches near-lossless image quality with an 80 $\sim$ 90\% reduction in computation requirements.



\bibliographystyle{IEEEtranN} 
\bibliography{paper}

%

\end{document}